# TESSELLATED BIOMES

*Distributed Robotic Assemblies for Architectural Resilience*

Sergio Mutis[1], Eric Hughes[2], Gert Duvenhage[3]
[1]*MIT, smutis@mit.edu, 0009-0006-4074-0734*
[2]*UCL, eric.hughes.21@ucl.ac.uk, 0009-0001-5447-950X*
[3]*MIT, duven616@mit.edu, 0009-0004-9451-0921*

**Abstract.** This paper presents *Tessellated Biomes*, a cyber-physical framework for the adaptive robotic construction and reconfiguration of modular multi-material assemblies. It challenges the linear lifecycle of standard construction by fusing (1) local microfactory fabrication, (2) discrete multi-material optimization, and (3) distributed robotic assembly into a unified circular process of spatial adaptation. The research details methods for the digital fabrication of self-aligning modular primitives in multiple materials (PLA, timber, and concrete) produced in local microfactories; the aggregation and optimization of these primitives into compression-based discrete structures; and the deployment of custom quadrupedal robots that collaboratively relocate material into realized physical aggregations. The framework is validated through the fabrication, optimization, and robotic assembly of discrete structures. Together, these results position *Tessellated Biomes* as a model for resilient, reconfigurable architecture.

## 1. Introduction

The contemporary construction industry remains constrained by linear lifecycles, rigid workflows, and resource-intensive processes that generate excessive waste and restrict adaptability. Buildings are typically fabricated through irreversible stages, contributing to a substantial environmental footprint: the sector consumes 32% of global energy and produces 34% of $CO_2$ emissions, with cement and steel alone responsible for nearly 18% (UNEP, 2024). These issues are exacerbated by the slow adoption of digital methods (RICS, 2023), limiting feedback between design, fabrication, and performance. A shift toward technologically enhanced circular systems is therefore pressing.

In contrast, social insects construct complex structures through collective, decentralized behaviours. Termite mounds regulate climate through continual material exchange (Caine et al., 2025), army ants form adaptive bridges by linking their bodies, and bees adjust comb geometry to structural and thermal demands.

These systems exemplify stigmergy, where environmental feedback drives self-organization without global plans (Boldini et al., 2024). Swarm robotics adopts similar principles, using local rules to coordinate large-scale behaviours (Salman et al., 2024). For architecture, these logics prompt a shift from top-down to bottom-up assembly that adapts to material availability and changing conditions.

## 2. State of the Art

### 2.1. HUMAN-ROBOT COLLABORATION (HRC)

Research on human–robot collaboration (HRC) in architecture has established the basis for cooperative fabrication environments where humans and machines share space, agency, and creative roles. Early work emphasized safety, precision, and co-presence, ensuring that robots could operate near humans without physical barriers (Bock and Linner, 2015; Gramazio and Kohler 2014). More recent studies highlight adaptability and the distribution of cognitive and creative agency between designers and robotic systems (Zhao et al., 2024; Skevaki et al., 2024), positioning the robot as an intelligent collaborator responding to gestures, sensing material feedback, and adjusting tasks in real time.

### 2.2. COLLECTIVE ROBOTIC CONSTRUCTION (CRC)

Collective robotic construction (CRC) extends HRC toward multi-agent systems capable of autonomous coordination. Instead of emphasizing co-presence, CRC distributes agency among robots that sense their context, communicate locally, and adapt, echoing biological systems such as termite colonies (Werfel et al., 2014; Smith et al.; Lutz, 2015). Subsequent work furthered sensing to help collectives negotiate stability and unstructured field conditions (Melenbrink et al., 2017), while later research coupled robot behaviour with material response (Weber et al., 2019). Learning-based approaches further enabled data-driven coordination as goals change (Hosmer et al., 2023, 2024; Mutis et al., 2023).

### 2.3. DISTRIBUTED NETWORKED FABRICATION

Distributed and networked fabrication challenges centralized production by enabling localized, digitally coordinated making (Carpo, 2023). The maker movement and early digital fabrication ecosystems (Gershenfeld, 2005) established community-embedded production models that decentralize fabrication beyond large industrial facilities. Contemporary practices such as AUAR and Beta Realities operate distributed fabrication networks where computation, material processing, and assembly are closely coupled. Parallel work researches micro-factories, digital maker communities, and globally connected yet locally enacted communities (Ünlü et al., 2025; Sanchez, 2020).

### 2.4. DISCRETE STRUCTURAL OPTIMIZATION

Discrete structures have become increasingly relevant to architectural computation, shifting from continuous surfaces to assemblies of finite, interlocking components. Early work showed how discrete units approximate topology-optimized distributions (Rossi and Tessmann, 2017), while digital materials (Popescu et al., 2006; Retsin, 2019; Retsin and Garcia, 2016) established combinatorial logics where performance emerges from part-to-part relations. Yet achieving structural robustness without compromising reconfigurability remains a major challenge, with stabilizing strategies such as post-tensioning or adhesives limiting reversibility.

### 2.5. GAPS & OPPORTUNITIES

CRC, distributed fabrication, and discrete structural optimization remain largely disconnected. CRC advances autonomous coordination but rarely considers fabrication constraints or multi-material behaviour; distributed micro-fabrication supports local production but is not structurally or robotically integrated; and discrete optimization seldom accounts for stock limits or reconfiguration. Building on this opportunity, we introduce Tessellated Biomes, an adaptive framework that links fabrication, optimization, and assembly.

## 3. Methodology

The Tessellated Biomes framework is organized into three interdependent components, which also operate as sequential steps in the system's workflow: (1) the local multi-material fabrication of modular self-aligning primitives, (2) the digital discrete structural optimization defining their spatial organization, and (3) the collaborative robotic reconfiguration of the primitives in physical space.

### 3.1. LOCAL MULTI-MATERIAL FABRICATION

Step 1 focuses on fabricating the modular primitives used for robotic assembly and structural interlocking. Each unit is a 5×5×5 cm block with 1.25 cm male–female indentations that provide vertical self-alignment and passive error correction during gripping. A local microfactory setup produces these components in PLA, timber, and concrete, each with distinct mechanical properties. The microfactory combines FDM printers, a CNC router, and modular casting stations within a compact 10–15 $m^2$ footprint, enabling small-scale, multi-material production on site.

(1) PLA components are fabricated on a standard Bambu P1S FDM printer (256 $mm^3$ build volume), which yields up to nine voxel units per batch. Elements longer than three voxels are produced as segmented male–female parts that interlock and are then bonded into continuous members (Fig. 1).

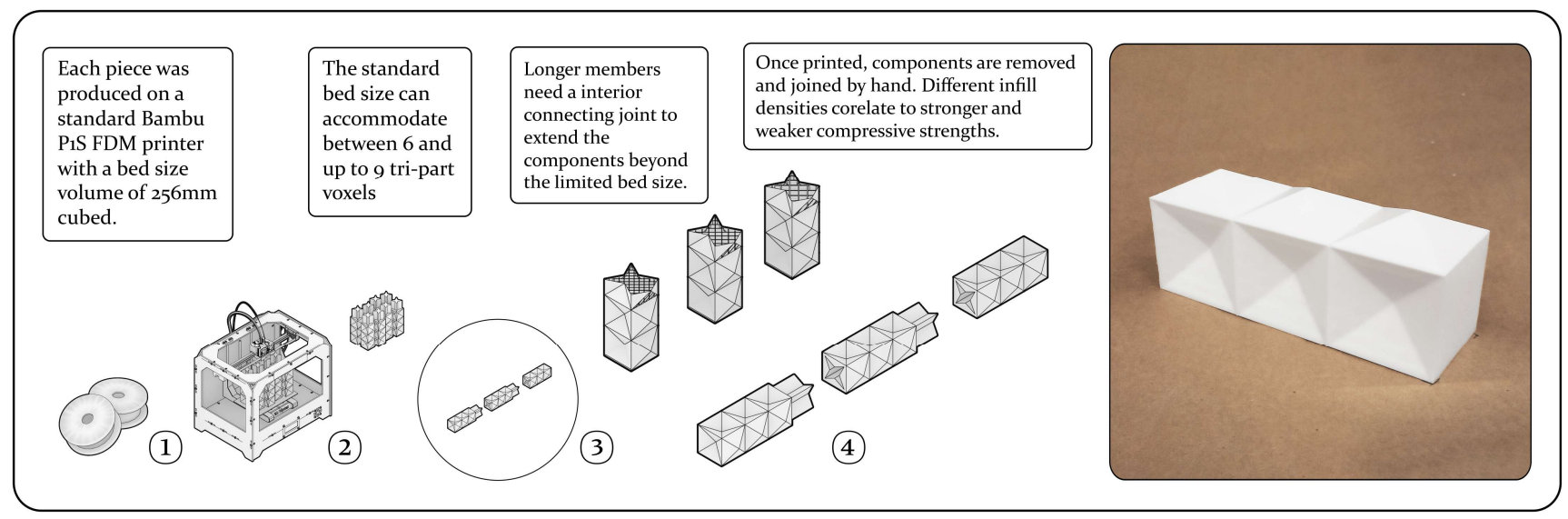

*Figure 1. PLA Voxel Fabrication Methodology*

(2) Timber components are milled from 50 mm red oak stock using a 48″ × 96″ jig that stabilizes the workpiece and ensures consistent cutting of the pyramidal self-alignment features on all faces. Joint locations are coordinated with the cutting sequence so the blade kerf does not remove essential geometry (Fig. 2).

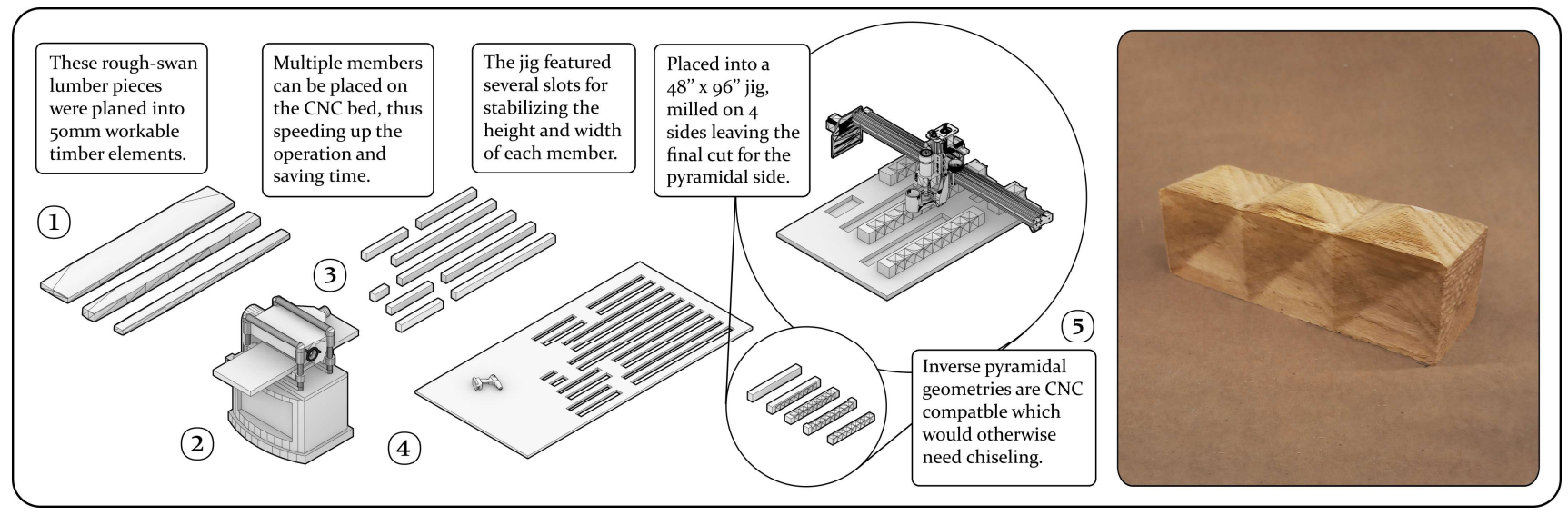

*Figure 2. Timber Voxel Fabrication Methodology*

(3) Concrete components are cast in modular four-sided PLA molds featuring a dovetail locking system and mitered edges for precise assembly and easy demolding. Mold length is adjustable, enabling both voxel-scale and extended elements to be cast within the same formwork system (Fig. 3).

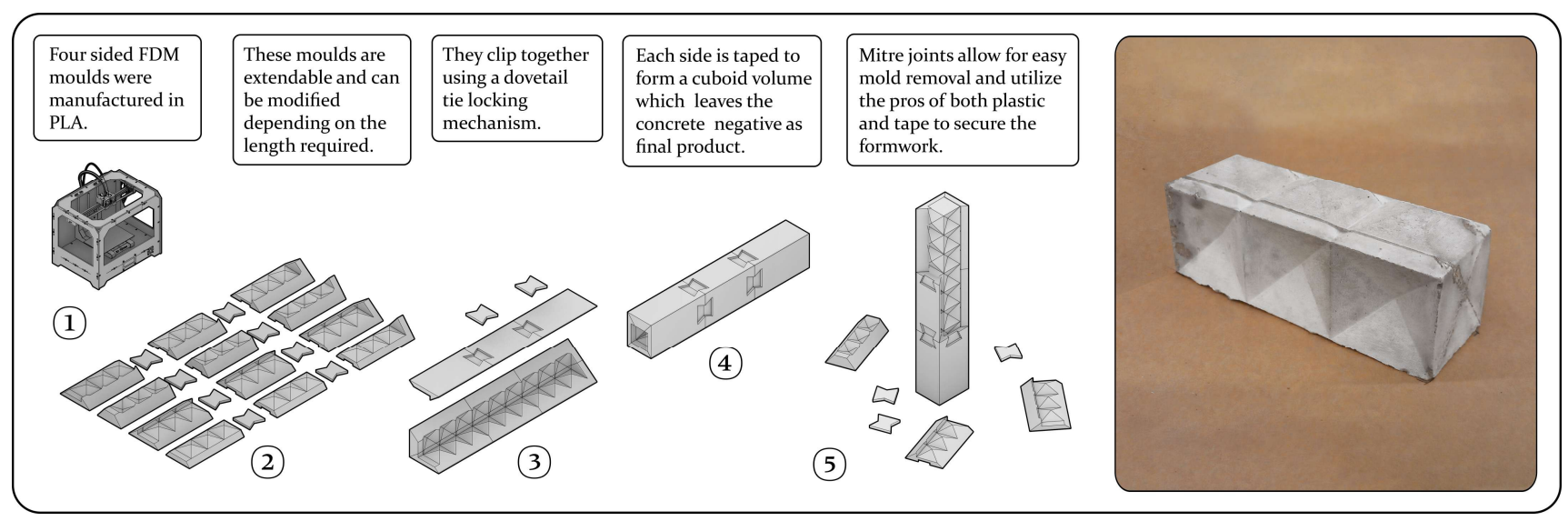

*Figure 3. Concrete/Rockite Voxel Fabrication Methodology*

## 3.2. DISCRETE STRUCTURAL OPTIMIZATION

Step 2 of the *Tessellated Biomes* framework employs a discrete structural optimization method leveraging the self-aligning components fabricated in Step 1. The goal is to identify assemblies that are stable, materially efficient, and robust. Four metrics structure this evaluation and optimization method:

(1) Compression Safety Index (CSI) measures the minimum ratio between compressive capacity and demand across all voxels: $CSI_p = (A_p\ f_c\ \phi) / N_p$, where $A_p$ is bearing area, $f_c$ material strength, $\phi$ a safety factor, & $N_p$ axial load.

(2) Volume Coverage (VC) quantifies the fill percentage filled under stock limits: $VC = N_{occ} / N_{total}$, where $N_{occ}$ is the # of occupied voxels.

(3) Structural Connectivity Index (SCI) measures how well components interlock by combining average contact degree $\bar{d}$, articulation ratio $r_{artic}$, and ground-connected fraction g: $SCI = \alpha \cdot \bar{d} + \beta \cdot (1 - r_{artic}) + \gamma \cdot g$.

(4) Total Weight (W) captures material usage and robotic payload: $W = \Sigma(\rho_p\ V_p)$, where $\rho_p$ and $V_p$ are the density and volume of each voxel.

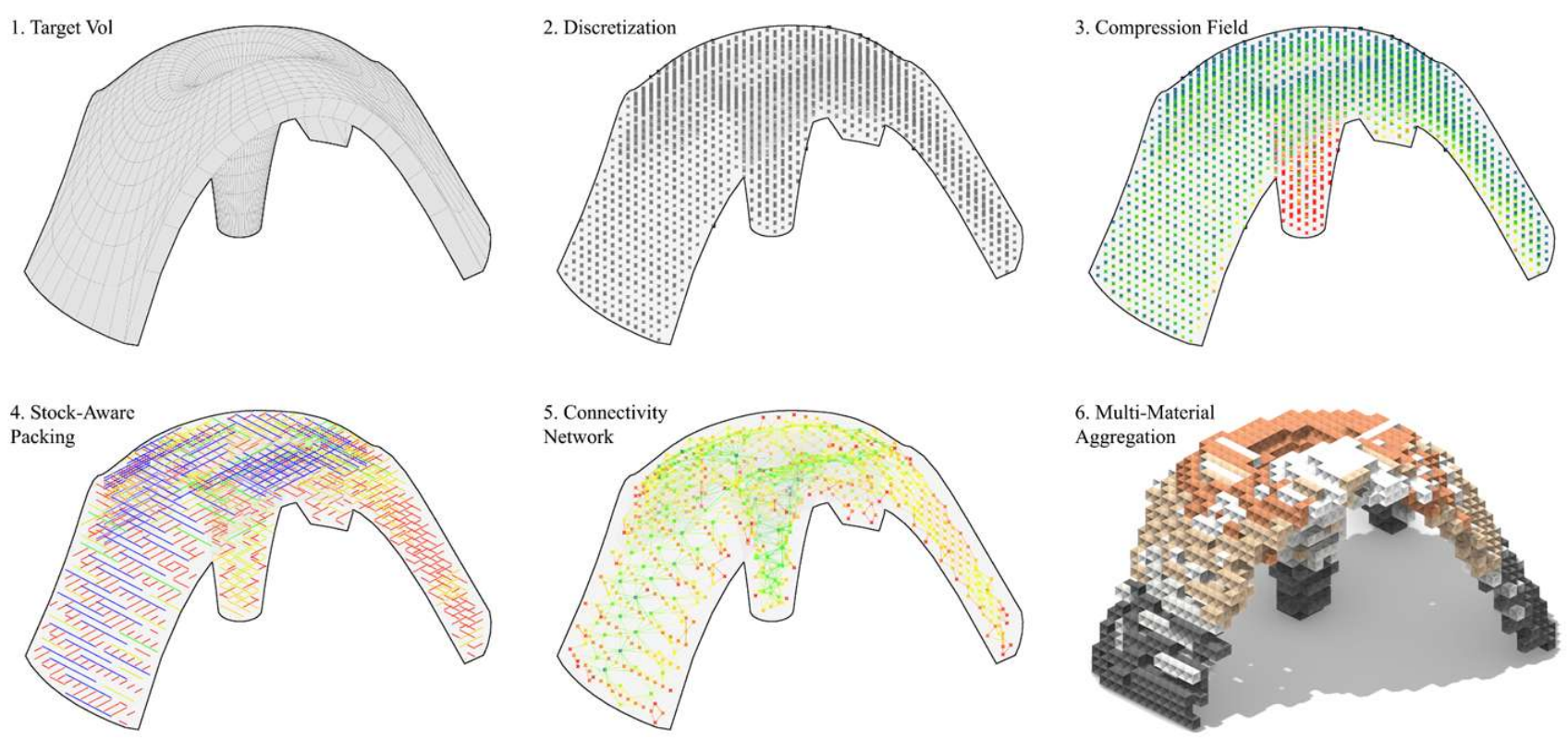

*Figure 4. Discrete Multi-Material Optimization Pipeline*

The optimization pipeline (Fig. 4) operates in six steps. (1) Target volumes are defined as compression-dominant geometries. (2) A uniform voxelization converts each volume into occupied voxel centroids. (3) A vertical compression field accumulates self-weight and applied loads down each column to yield voxel-level demand. (4) A rule-based material assignment selects PLA, timber, concrete, or higher-strength fillers with sufficient capacity. (5) A stock-aware packing algorithm assembles components bottom-up from a CSV inventory, enforcing discrete counts and alternating layer orientations. (6) A connectivity graph is generated, from which CSI, coverage, SCI, and weight are evaluated.

### 3.3. COLLABORATIVE ROBOTIC RECONFIGURATION

*Figure 5. Quadrupedal Robot Recolating Material Primitive.*

Step 3 of the *Tessellated Biomes* framework involves the assembly of the material primitives fabricated in Step 1 into the optimized compression structures calculated in Step 2 by a group of custom quadrupedal robots.

Each robot (Fig. 5), measuring 37.6 × 37.6 × 22.2 cm in its natural stance and weighing 1.45 kg, consists of a central body and four articulated legs. The central body houses a Raspberry Pi 2 as the onboard computer, a U2D2 interface and power hub for motor control and communication, a central lower gripper actuated by two AX-12A Dynamixel servo motors, and a 12 V / 5 V 3000 mAh Li-ion battery providing power. Each leg offers three degrees of freedom, driven by three additional AX-12A servos daisy-chained via 3-pin TTL.

The gripper teeth and foot tips match the voxel indentations, sliding into position at each grip or step. With 5×5 cm modules and central indentations, placement errors under 2.5 cm are passively corrected. This geometry-guided alignment simplifies catching, navigation, and placement, removing the need for external sensing or visual tracking.

A cyber-physical control system synchronizes the robots with a real-time simulation environment. A central computer running the *Tessellated Biomes* simulator communicates via Wi-Fi with each Raspberry Pi onboard, which relays motor commands through the U2D2 interface to the AX-12A servos daisy-chained inside the robot. Motion trajectories and joint angles are computed in simulation and executed physically, while the motors feedback position and load data to maintain synchronization. This closed-loop enables behaviours to be tested virtually before deployment and ensures cyber-physical synchronization.

## 4. **Results**

### 4.1. MICROFACTORY MULTI-MATERIAL FABRICATION

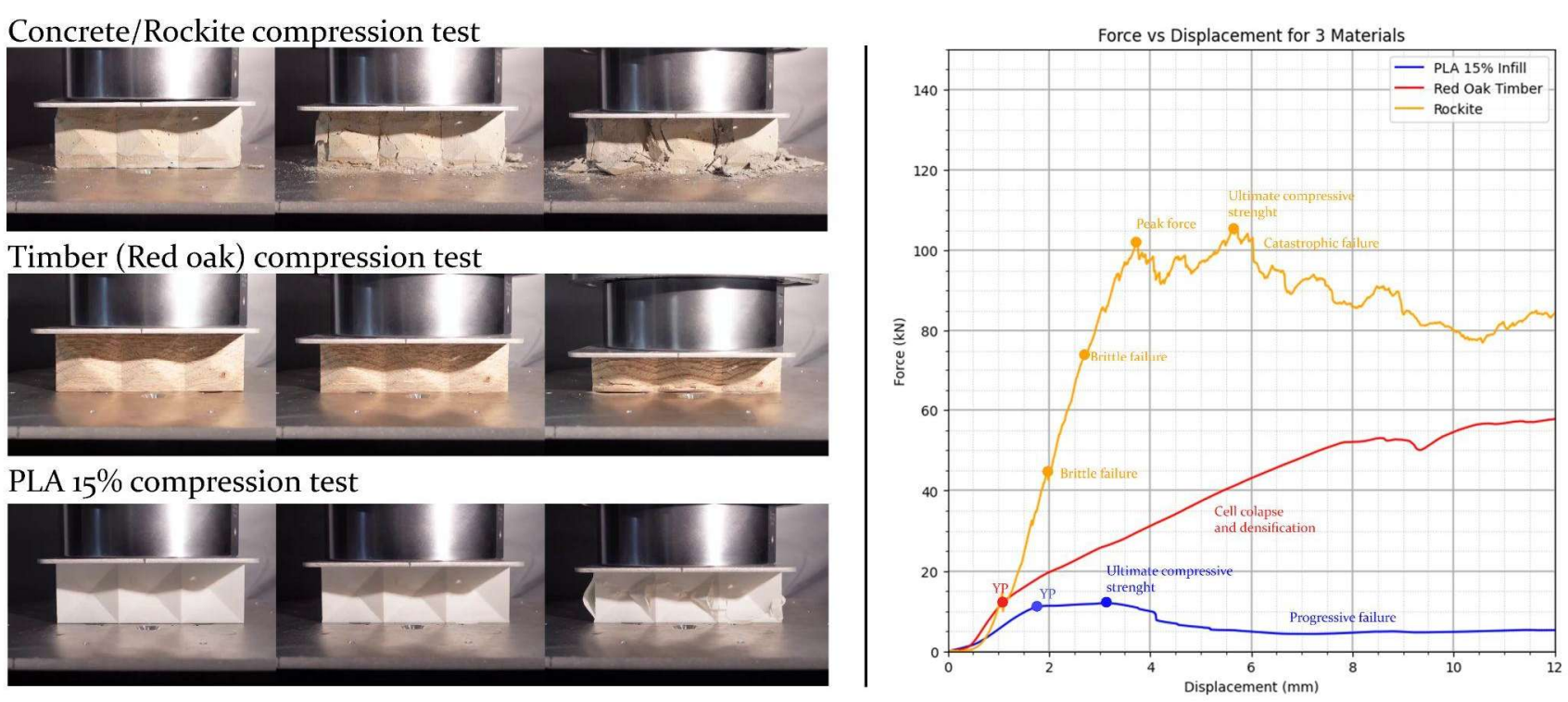

*Figure 6. Material Compression Testing Images and Results*

Across the three microfactory streams, we produced a total of 1,420 voxels, corresponding to 1198 PLA voxels, 126 red oak voxels, and 96 concrete/Rockite voxels. Given the average primitive length of 7-7.5 voxels, this stock translates to approximately 160 PLA, 20 timber, and 15 concrete components, providing the material inventory for subsequent optimization and robotic assembly.

To assess structural viability, representative units from all three materials were tested in compression using an Instron universal testing machine (Fig .6). Because the voxels include pyramidal alignment features, the initial contact area was smaller than the full 5×5 cm footprint and expanded during deformation, yielding conservative capacity readings.

The tests revealed clear performance stratification across materials. Concrete/Rockite achieved the highest capacity, failing at approximately 100 kN after 4 mm displacement. Red oak exhibited a micro-failure pattern, reaching ~54 kN at 9 mm displacement without catastrophic collapse. PLA displayed the lowest capacity, failing at ~12 kN at 2 mm displacement. Despite these differences, all materials exceeded the required load thresholds for the discrete compression structures used in the subsequent optimization and robotic assembly stages. These results confirm that the microfactory-fabricated primitives, across plastics, timber, and cast composites, are structurally suitable for use in modular, compression-dominant assemblies.

## 4.2. DISCRETE ASSEMBLY OPTIMIZATION

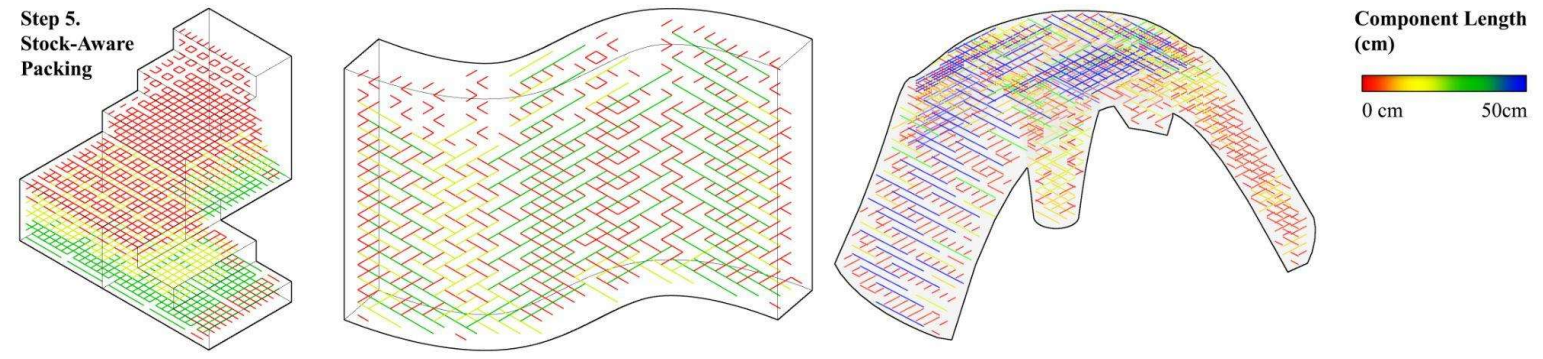

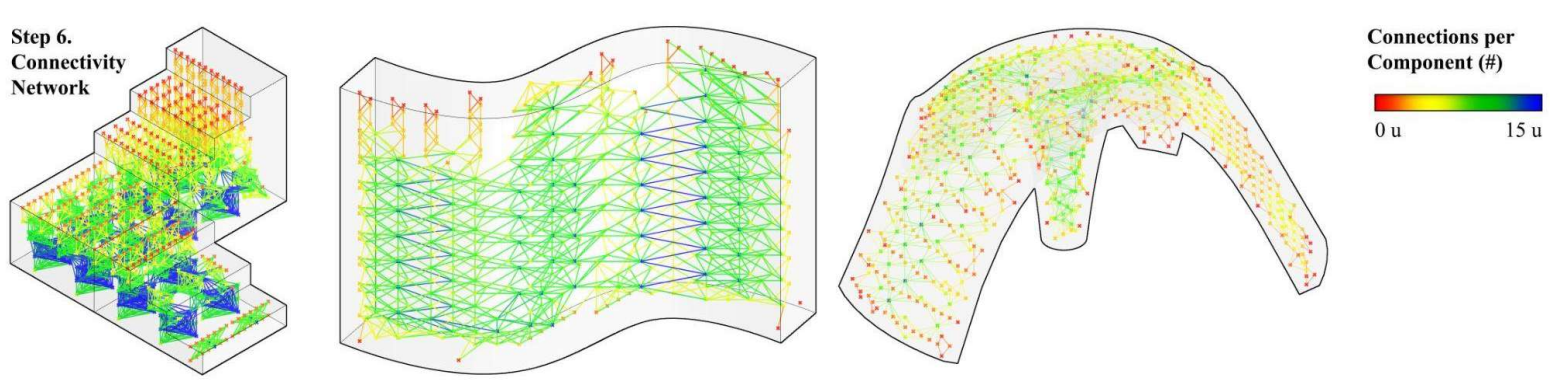

*Figure 7. Sample Keyframes of Optimization Pipeline for Stairs, Wall, and Shell*

We evaluated the Tessellated Biomes optimization method across three typologically distinct compression-dominant volumes (staircase, curved wall, and shell) chosen to test performance under varying curvature, thickness, and vertical load paths. Each geometry was voxelized at 5 cm resolution, and evaluated under three material stocks (S1: PLA-heavy, S2: timber-dominant, S3: mixed PLA–timber–concrete). Across all testing, the evaluation followed the same sequence: voxelization, compression field, material assignment, packing under stock constraints, and connectivity analysis (Fig.7).

Results show that geometry and stock composition strongly influence performance across the three test volumes. The staircase achieved the highest CSI due to short vertical load paths, while the shell benefited most from mixed-material stocks that enabled lateral thickening. The wall produced the greatest SCI because of continuous ground contact and broad planar interfaces. Volume coverage varied with inventory limits, and total weight tracked material density, with timber-dominant stocks yielding lighter assemblies better suited for robotic handling. Overall, the pipeline generates structurally legible, materially feasible assemblies while clarifying how stock configurations shape safety, connectivity, weight, and coverage.

## 4.3. COLLABORATIVE ROBOTIC RECONFIGURATION

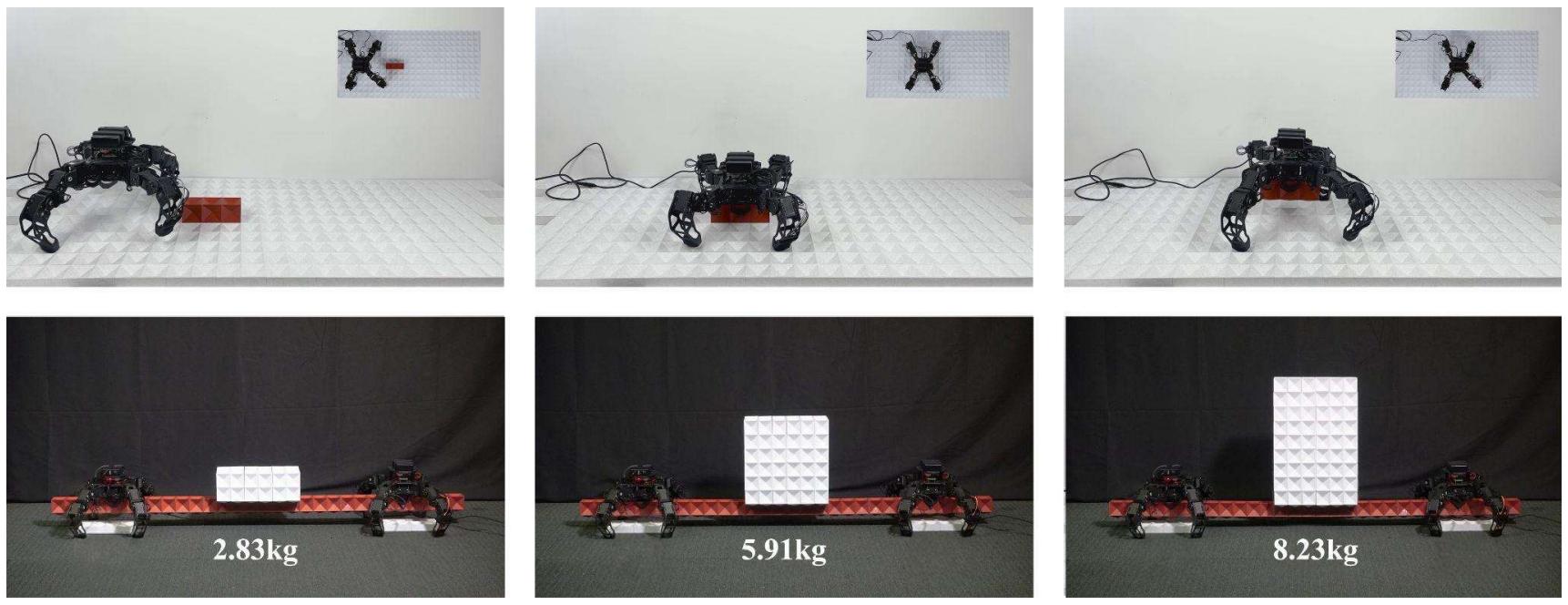

*Figure 8. Behaviour Sequences: Locomotion (top) Collaborative Lifting (Bottom)*

The quadrupedal robots demonstrated a repertoire of behaviours required for distributed material reconfiguration (Fig. 8). Agents achieved stable planar locomotion and reliable gripping, lifting, and transport of voxel units. They climbed slopes up to 45° before losing traction, and lifted loads up to 2.83 kg individually; in pairs, robots collaboratively transported components up to 8.23 kg, 2.84× their body weight. Failures were caused by geometric imbalance rather than motor limits. Robots also carried elongated elements up to 1.5 m, 3.99× their length, with no observed stability issues in cooperative transport. Together, these behaviours validate geometry-guided gripping and passive alignment as effective substitutes for vision-based perception in multi-robot handling.

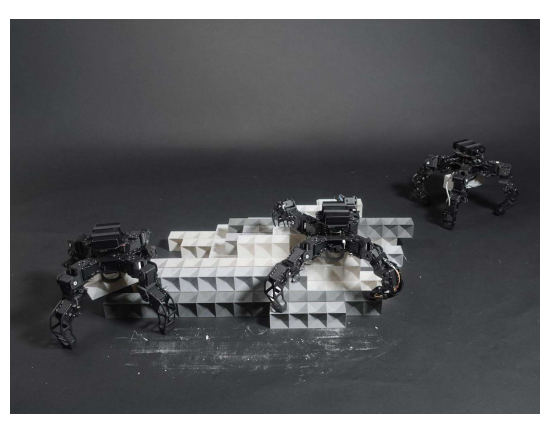
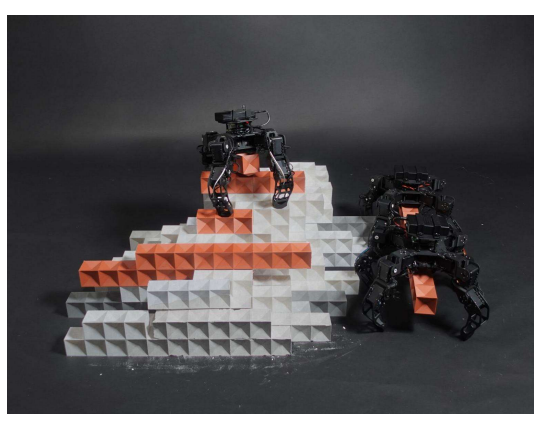
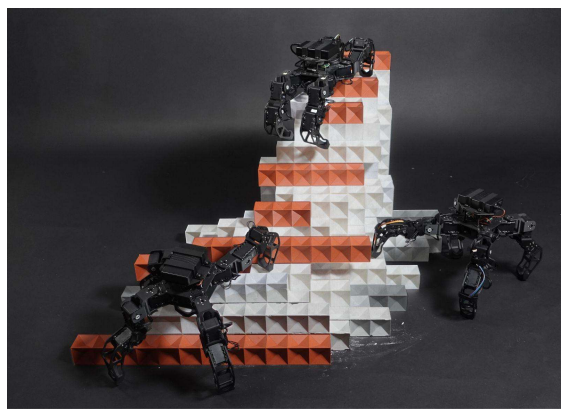

*Figure 9. Stair Reconfiguration Sequence With 3 Robots and 40 Material Components*

Building on these capabilities, teams of three robots assembled simple walls, stepped configurations, and columns between 0.60 m and 1.40 m high using ~100 available components (Fig. 9). Assemblies adhered to stock constraints defined in the optimization step, demonstrating tight correspondence between digital packing decisions and physical execution. Across all trials, the self-aligning indentation geometry ensured consistent placement accuracy without external sensing or tracking. These results confirm that the established behavioural set, locomotion, gripping, lifting, collaborative transport, and inclined climbing, supports the physical realization of optimized assemblies within the *Tessellated Biomes* framework.

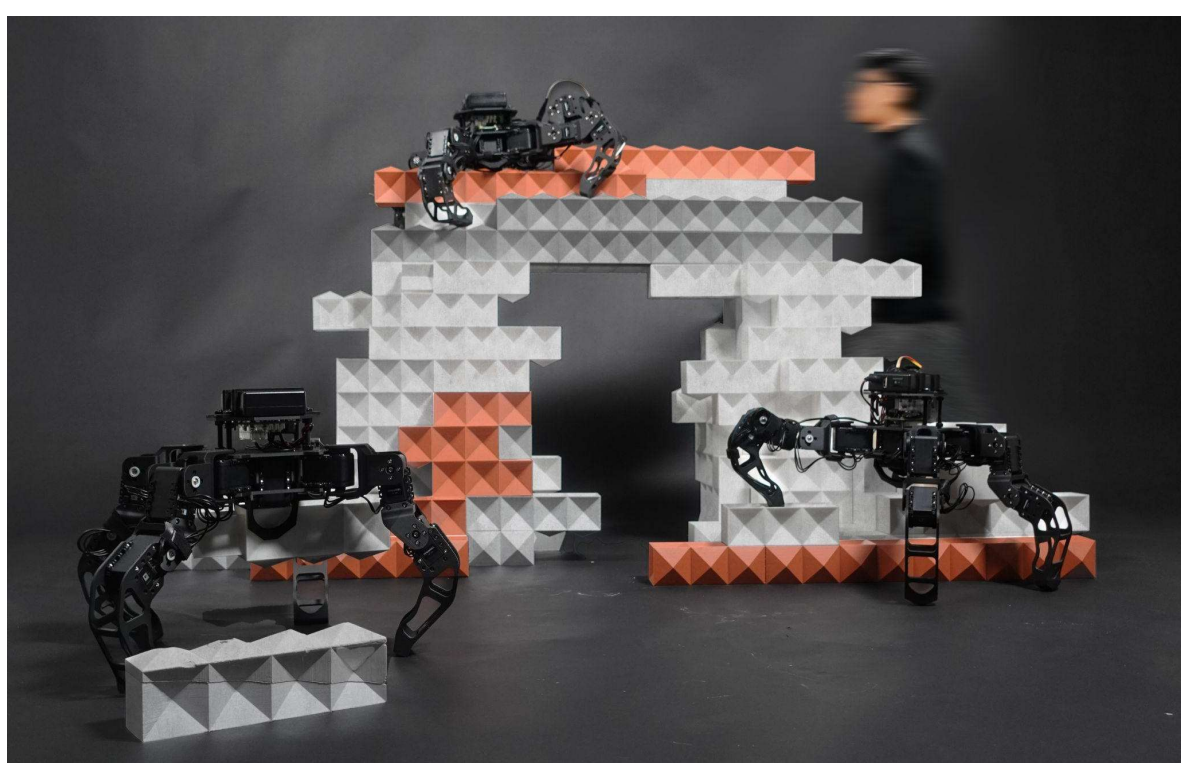

*Figure 10. Custom Robots Navigating a Medium-Scale Discrete Multi-Material Assembly*

## 5. Discussion & Conclusion

*Tessellated Biomes* demonstrates how local multi-material fabrication, discrete stock-aware optimization, and collective robotic assembly can operate as an integrated adaptive construction framework. Each subsystem was validated through methods suited to its domain: multi-material microfactories were assessed through production output and compression testing, the optimization pipeline through three compression-dominant digital prototypes, and the robots through a catalogue of physical behaviours and multi-agent assembly trials. Together, these results indicate that distributed robotic construction can be grounded in real materials, limited inventories, and minimal sensing while retaining the capacity for reconfiguration.

The findings also reveal limitations that point to clear areas for refinement. Compression tests used flat Instron platens against pyramidal interfaces; geometry-matched fixtures would improve accuracy. Repeated robotic handling caused wear, particularly in cast components, indicating the need for more durable mixes. Structurally, the model omits friction and reversible joints, and full-scale behaviour remains untested. Robotic constraints in payload and reach likewise limit achievable assemblies. Collectively, these limitations identify where material, structural, and robotic performance can be strengthened.

Future work includes friction-aware structural modelling, reversible or active joints, and larger-scale testing; robot-level learning for gait adaptation, collaborative transport, and sequence planning; and system-level feedback loops linking fabrication, optimization, and assembly in near real time.

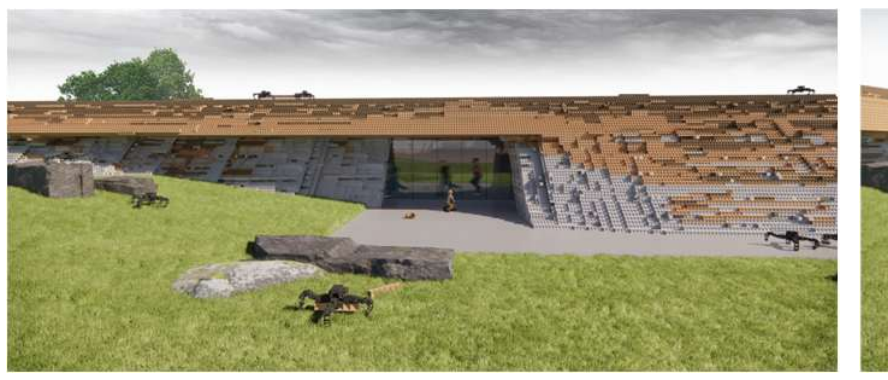
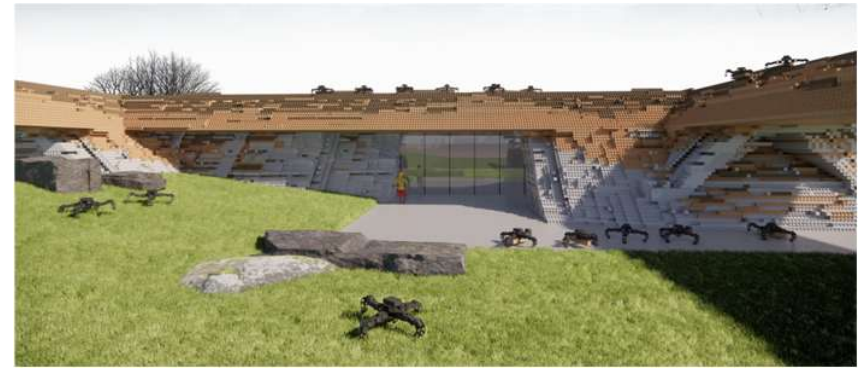

*Figure 11. Renders - Vision of the Evolution of a Tessellated Biome over Time.*

*Tessellated Biomes* ultimately positions architecture as a reconfigurable, material-aware, and continuously adaptive system. By linking distributed fabrication, multi-material optimization, and collective robotic assembly, it offers a path toward circular, resilient spatial ecologies capable of evolving rather than remaining fixed.

**Acknowledgements**

We thank the *Living Architecture Lab* at University College London (UCL), led by Tyson Hosmer, Octavian Gheorghiu, and Philip Sieddler, for providing the intellectual lineage that informed this research. We also thank the staff of MIT's *How to Make (Almost) Anything* course for supporting early prototyping efforts. We are grateful to Skylar Tibbits and Daniela Rus for their guidance, and to Christopher Dewart for his technical assistance in the fabrication of the modular primitives.